\documentclass[10pt,twocolumn]{article}
\pdfoutput=1
\usepackage{iccv}
\usepackage{times}
\usepackage{epsfig}
\usepackage{graphicx}
\usepackage{amsmath}
\usepackage{amssymb}
\usepackage{bbm}

\usepackage[pagebackref=true,breaklinks=true,letterpaper=true,colorlinks,bookmarks=false]{hyperref}

\newcommand\indep{\protect\mathpalette{\protect\independenT}{\perp}}
\def\independenT#1#2{\mathrel{\rlap{$#1#2$}\mkern2mu{#1#2}}}

\iccvfinalcopy 

\def\iccvPaperID{****} 

\ificcvfinal\fi
\begin{document}

\title{Explaining Visual Models by Causal Attribution}

\author{\'Alvaro Parafita, Jordi Vitri\`a\\
Universitat de Barcelona\\
{\tt\small \{parafita.alvaro,jordi.vitria\}@ub.edu}
}

\maketitle

\begin{abstract}
   Model explanations based on pure observational data cannot compute the effects of features reliably, due to their inability to estimate how each factor alteration could affect the rest. We argue that explanations should be based on the causal model of the data and the derived intervened causal models, that represent the data distribution subject to interventions. With these models, we can compute counterfactuals, new samples that will inform us how the model reacts to feature changes on our input. We propose a novel explanation methodology based on Causal Counterfactuals and identify the limitations of current Image Generative Models in their application to counterfactual creation.
\end{abstract}

\section{Introduction}

The field of Machine Learning (ML) Interpretability has been gaining traction over recent years due to the proliferation of ML applications in a wide range of uses. However, there is no common notion of what interpretability means, or what does it entail. Lipton \cite{lipton_mythos_2016} distinguishes two types of interpretability: transparency and post-hoc explanations. We will focus on the latter, specifically on the definition of a system capable of attributing the effect of feature alterations to the outcome of a predictor. Although the technique is general (from regression to classification; from tabular data to image data), in this paper we will only cover its application to Deep Learning image classifiers.

As an example, consider a CNN image classifier that tries to predict the gender of a person from a face. Our objective is not to optimize the classifier, but to explain what affects its prediction for specific faces. Assuming we know certain factors about the face (\ie hair colour, age, use of makeup or presence of facial hair), we are interested in detecting which of these latent factors (variables that describe the image but which are not included in the input) are responsible for the outcome of the classifier. This contrasts with the problem of saliency, where pixels are sorted by importance with respect to the result of the prediction. Here we attribute more semantically-charged aspects of the input, its latent factors, since those are the ones that can be interpreted and understood by the human user. Specifically, we will answer questions of the type:

\begin{quote}
Given the fact that this face belongs to a woman, has been classified as a woman and the person does not have a beard, how would the classifier's prediction have changed had there been a beard?
\end{quote}

\begin{figure}
    \centering
    \includegraphics{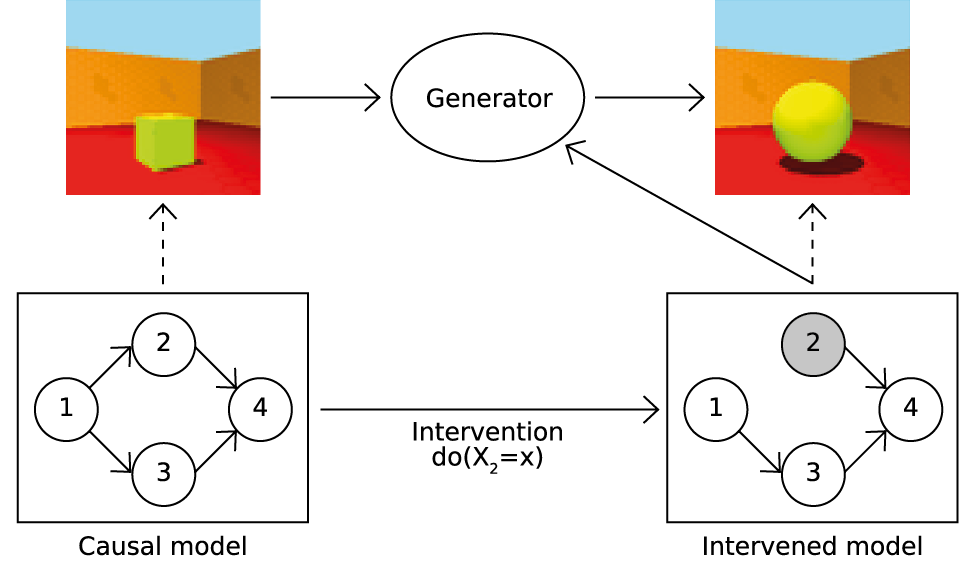}
    \caption{Explanation of visual models by counterfactuals needs to consider the causal structure of the data in terms of meaningful latent factors. It is by applying interventions on this causal model that we create a new, intervened model on which we can base generators to create causal counterfactual images. In this example, by intervening on shape, scale increases as well.}
    \label{fig:explanation}
\end{figure}

In the field of Causality (refer to the book by Pearl \cite{pearl_causality}), this is referred to as Counterfactual reasoning. The idea is to consider alternative worlds where certain aspects have changed \textit{independently of their underlying causes} (this is what we call an \textit{intervention}), but the rest of the variables will be unaltered except for those that are affected by the intervention we made. This is a Causal Counterfactual.

We find that these contrastive questions are the key to explaining the effects of factors on a predictor. Figure \ref{fig:explanation} summarizes our views on the importance of causal models for the computation of counterfactuals. Assume we have an image to explain, along with its corresponding latent factors. We want to compute the effects of certain interventions (modifications on some of the latent factors) on the classifier prediction. These interventions alter the causal model in which we work: from the original one, that represents our dataset, to the intervened one, that contains the desired intervention. Counterfactuals are, in fact, samples from that intervened model adjusted to the input we want to study. By taking several of these samples (new latent factors) we can generate, through a Counterfactual Image Generator, new images that contain the intervention we made, along with any other factors affected by this intervention. Finally, these modified (counterfactual) images pass through the classifier to obtain different predictions, and by aggregating them we obtain the average effect of the intervention on the classifier, for the original input image that we want to explain.

Notice that the original causal model and the intervened one may result in significantly different distributions.  Therefore, only by considering the causal graph can we really sample from the intervened model, where counterfactuals live. A non-causal generator would sample from the observational data, represented by the original causal model, but it would not accommodate for the causal effects of our interventions.

The computation of effects of interventions on the final predictor allow us to explain why a certain prediction was made. If we notice that by intervening on a factor the prediction changes significantly, we can say that the current value for that factor is a cause for the current prediction. Not only that, but we also know what could change in order to alter the outcome of the classifier. Additionally, this technique can be applied to detect classifier biases or unfairness in its assessment (\ie, race as a factor in recidivism prediction).

Lastly, for clarification, it is important to note that the term \textit{counterfactual} has been used with very different meanings across the literature. As an example, Sundarajan \etal \cite{sundararajan_gradients_2016} used it to refer to interpolations between an image and a baseline (a blank picture), as a way to estimate the gradient of the classifier in non-saturated regions of the input. Wachter \etal \cite {wachter_counterfactual_2017} and Goyal \etal \cite{goyal_counterfactual_2019}, on the other hand, consider a counterfactual as an image similar to the original one but resulting on a different prediction. The difference with the definition in Causal Theory is subtle, but decisive: causal counterfactuals stem from an intervention on the factors of the causal model, whilst Wachter and Goyal's result from looking for a different outcome of the predictor, and only then do they look at the factors. Their use of those counterfactuals is to compare between the original input and the new one, to highlight the differences between them as the reasons for the change in classification. This, however, does not take into account that some changes might appear in order to make the true causal factor more likely, \textit{confounding} the real causal factor with a non-causal one. Only by intervention can we really estimate the effect of a factor. 

Our contributions are threefold:

\begin{enumerate}
    \item We define a method for implementing Causal Graphs with Deep Learning that allows sampling from the observational model and any intervened model. Additionally, we can also estimate the log-likelihood of any sample. Both operations allow backpropagation, so they can be used as parts of even more complex models. As far as we know, this approach is novel in Deep Learning; we call this method Distributional Causal Graphs.
    \item We define a new explanation technique for visual models based on latent factors, using the previously defined graphs and a Counterfactual Image Generator.
    \item We identify the limitations of current Conditional Image Generators for Counterfactual Generation and enumerate the requirements for such a model.
\end{enumerate}

\section{Related work}

The use of contrastive techniques for attribution (comparing the image of study to another with certain features changed) is well extended in the literature. Zeiler \etal \cite{zeiler_visualizing_2014} and Ancona \etal \cite{ancona_towards_2018} use pixel masking (replacing certain pixels with constant, blank values) to measure classifier dependency on certain pixels. In this case, they compare the original image with a masked one. Chang \etal \cite{chang_explaining_2018} also use pixel masking, not only by replacing with a constant value, random noise or by blurring the masked region, but by in-filling it with probable values with respect to the remaining parts of the image. This in-filling is done via Generative Models.

Counterfactuals understood in the sense of minimal alterations to change classifier prediction are used in the aforementioned Wachter \etal \cite{wachter_counterfactual_2017} and Goyal \etal \cite{goyal_counterfactual_2019}, but also in Mothilal \etal \cite{mothilal_explaining_2019}. This latter case also mentions the importance of counterfactual diversity (obtaining diverse counterfactual images that manage to change classifier prediction) and counterfactual feasibility (likelihood that such a counterfactual exists or that the intervention necessary is feasible to execute).

Parallel work by Denton \etal \cite{denton_detecting_2019} conceives counterfactuals in the sense of altering just one of the latent factors and estimating its effect. However, their work does not take into account the causal dependencies resulting from this transformation, and the use of their technique is limited to interventions of single variables. Their approach is similar to TCAV (Kim \etal \cite{kim_interpretability_2017}): computing vectors that encode each feature of interest and moving along this direction to study the effects on the classifier. Although the latter uses them in the space of layer activations inside the classifier, the former works in the latent space of a Generative Adversarial Network (Goodfellow \etal  \cite{goodfellow_generative_2014}), so that the changes produced by the explanation technique still result in realistic images.

In terms of Counterfactual Image Generators, CausalGAN  \cite{kocaoglu_causalgan:_2017} uses a Causal Controller to inform a Conditional GAN of which latent factor configurations to use in the generation of new images. Our Causal Graph is based on their Causal Controller with a redefinition to allow for the computation of the log-likelihood of any latent factor sample. Their generator, however, is not sufficient for the computation of effects of interventions, as discussed in the following sections. We will also comment on Fader Networks (Lample \etal \cite{lample_fader_2017}) and AttGAN (He \etal \cite{he_attgan:_2019}) as Conditional Image Generators, and their limitations in the application to Counterfactual Image Generation.

\section{Background}

We start our exposition with a brief introduction to the framework of causality focused on our specific application.

Consider a set of random variables $\mathcal{V} = \{X_1, \dots, X_n\}$ and a set of exogenous random variables $\mathcal{E} = \{E_1, \dots, E_n\}$. We assume that the variables in $\mathcal{E}$ are mutually independent and also independent to any variable in $\mathcal{V}$ other than their corresponding one, meaning $\forall i \neq j, E_i \indep E_j, \, E_i \indep X_j$ but $E_i \not\indep X_i$. We define a probability distribution over the exogenous variables, $P_\mathcal{E}(e)$, while $\mathcal{V}$ is defined by a set of deterministic functions $\mathcal{F} = (f_1, \dots, f_n)$, one for each of the variables in $\mathcal{V}$. Each function $f_i$ relates their corresponding variable $X_i$ to its \textit{causes} (other variables in $\mathcal{V}$) and the corresponding exogenous noise variable $E_i$. Therefore, if $Pa_i, \: \emptyset \subseteq Pa_i \subsetneq \mathcal{V}$, denotes the set of causes of $X_i$, $X_i = f_i(Pa_i, E_i)$. Therefore, $f_i$ is a deterministic function relating the value of variable $X_i$ to the value of each of its causes $Pa_i$, plus an independent, exogenous noise variable $E_i$ that accounts for the stochasticity of the relationship. Altogether, this framework allows us to define what Pearl \cite{pearl_causality} calls a Structural Causal Model $\mathcal{M} = (\mathcal{V}, \mathcal{E}, \mathcal{F}, P_\mathcal{E})$. 

Note that the set of functions $\mathcal{F}$ defines a directed graph $\mathcal{G}=(\mathcal{V} \cup \mathcal{E}, E)$ where $E=\{(V_j, V_i) \mid \forall i = 1..n, \: \forall j : V_j \in Pa_i\} \cup \{(E_i, V_i) \mid \forall i = 1..n\}$. The set of edges that converge in a node defines the causal dependencies of that node ($Pa_i$) and the stochastic component of that relationship ($E_i$). We will only consider graphs with no cycles; therefore, we study Structural Causal Models whose implicitly defined graphs $\mathcal{G}$ are Directed Acyclic Graphs (DAG). For notational simplicity, we assume that nodes are sorted in a way that all parents of a node must have a lower index than that node. Since the graph is a DAG, a topological ordering of the graph induces such an index order. Therefore, $\forall i \neq j, X_j \in Pa_i \rightarrow j < i$.

\begin{figure}
    \centering
    \includegraphics{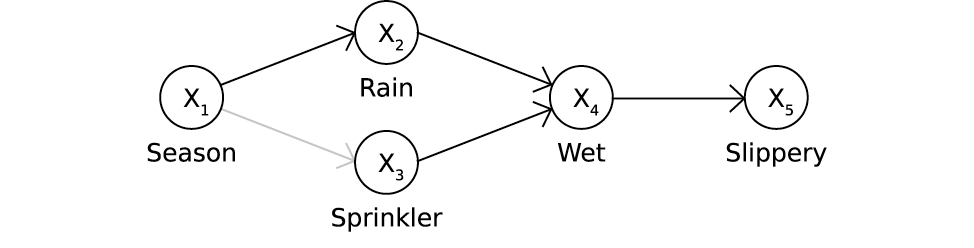}
    \caption{Wet floor example, proposed in \cite{pearl_causality}. The gray arrow represents the one that disappears when intervening on Sprinkler, creating the new intervened model.}
    \label{fig:wet_example}
\end{figure}

As an example, consider the graph in figure \ref{fig:wet_example}. We are interested in determining when the floor will be slippery ($X_5$). We know that the floor is slippery when it is wet ($X_4$), and it can be wet either because there has been rain ($X_2$) or because our neighbours' sprinkler has been activated ($X_3$). Finally, we know that rain probability depends on the current season ($X_1$), and that our neighbours only program their sprinkler in non-rainy seasons. Therefore, we can infer causal relationships between our variables, such as the ones depicted in the graph in Figure \ref{fig:wet_example}. Note that we omit the exogenous set of noise variables $\mathcal{E}$ from graph visualizations because there are no edges between them and other nodes other than their corresponding ones, so their presence is implicit.

Interventions in causal theory mean fixing the values of certain nodes (independently of the value of their parents) and perform predictions on the remainder of the graph. Interventions can be understood as removing any edges leading to the intervened nodes (both from variables in $\mathcal{V}$ and $\mathcal{E}$) and specifying the value that node ought to take. This allows to ask questions such as: "will the floor be slippery if we turn off the sprinkler?". Interventions are denoted as $\mathit{do}(X = x)$. In our example, the intervention is $\mathit{do}(\mathit{sprinkler} = \mathit{off})$.

Counterfactuals, on the other hand, are always based on a specific sample and an intervention to apply. Given both, we fix the noise signals ($\mathcal{E}$) according to the sample, perform the given intervention and run predictions on the resulting model. This allows to ask questions such as: "given that we know that the floor was slippery, the season is summer, it has not rained today and the sprinkler was on, would the floor still have been slippery had we turned the sprinkler off?". In this case, we perform the same intervention but the noise signals are conditioned on the fact that we saw the initial given sample. In the counterfactual world (the model with the applied intervention and the affected noise signals), rain is independent to the sprinkler state and it has not rained, therefore, rain cannot have started in any counterfactuals. Since the sprinkler was not activated and it has not rained, the floor cannot be wet and, consequently, it cannot be slippery either. 

The intervened causal model we will use to sample counterfactuals can be defined from modified sets $\widehat{\mathcal{F}}$ and $P_{\widehat{\mathcal{E}}}$. Given the original $P_\mathcal{E}(E_1, \dots, E_n)$ and the input on which we base the counterfactual, $x = (x_1, \dots, x_n)$, we condition the noise variables $\mathcal{E} = (E_1, \dots, E_n)$ on the current observation $x$. Therefore, the counterfactual model has an exogenous variable distribution $P_{\widehat{\mathcal{E}}}(e) = P_\mathcal{E}(e \mid x)$. Now, given an intervention $\mathit{do}(X_i = x'_i, \: \forall i \in I)$, the set of deterministic functions $\mathcal{F}$ is replaced by $\widehat{\mathcal{F}} = \{f_i \in \mathcal{F} \mid \forall i \not \in I\} \cup \{\widehat{f_i} := (X_i = x'_i) \mid \forall i \in I \}$. This replaces any functions that defined the intervened variables with a simple assignment to the intervention value.

Therefore, given a sample $x$ and an intervention $\mathit{do}(X_i = x'_i, \: \forall i \in I)$, counterfactual inference means computing predictions on a new Structural Causal Model $\widehat{\mathcal{M}} = (\mathcal{V}, \mathcal{E}, \widehat{\mathcal{F}}, P_{\widehat{\mathcal{E}}})$ with $\widehat{\mathcal{F}}, P_{\widehat{\mathcal{E}}}$ defined as above. Figure \ref{fig:wet_example} shows the intervened model if we omit the gray edge, which points to the intervened variable Sprinkler. In that setup, Sprinkler becomes independent of Rain, which was not true in the original causal model, which reflected our observational data.

\section{Proposed method}

This section is divided in three subsections, corresponding to the different parts of our attribution method. We begin with our Distributional Causal Graph, which will allow us to sample and compute likelihoods of the samples. Afterwards, we will cover the requirements for a Counterfactual Image Generator and the limitations of current approaches. Finally, we will detail the process through which we can estimate the effects of interventions and, therein, compute the attribution of latent factors.

\subsection{Distributional Causal Graph (DCG)}

Our objective is to explain predictions of an image classifier, but these explanations will be based on latent factors, meaning, a set of descriptive random variables (both discrete and continuous) that express semantically-charged aspects of the input. We assume that these factors are known for all instances in our training data, either through a classifier that predicts them or by being present in the dataset. They are interdependent in a causal way and we know a causal graph that describes their relationship. As said before, we omit any possible cycles in their causal relationships, hence the graph must be a DAG.

Our approach to modeling this graph is based on the Causal Controller from Kocaoglu \etal \cite{kocaoglu_causalgan:_2017}: each node is represented by a Multilayer Perceptron (MLP). The input of this MLP is all of its parent values (if any) concatenated as consecutive vector dimensions, with the additional concatenation of the corresponding exogenous noise variable. In the case of Kocaoglu's Causal Controller, the output of the MLP is a sample of the latent factor. Our DCG diverges from their implementation in terms of input and output of each node.

First, we need to make additional assumptions about the data in order to proceed. In the example from Figure \ref{fig:wet_example}, all variables are discrete, binary variables. All of them can be modelled as Bernoulli distributions, except for $X_1$, \textit{Season}, which is a 4-dimensional Categorical distribution. Then, each of our MLP outputs will be an estimation of the distribution parameters of the corresponding node, depending only (using as input) on its parent node samples. In the example, this means that all nodes except for $X_1$ will be represented by an MLP that takes its parent values and returns the parameter $p$ that represents the probability of the node achieving value 1, under the current circumstances. $X_1$, on the other hand, will output 4 different probability parameters, $(p_j)_{j=1..4}$, that represent the probability of each possible outcome (each of the seasons). 

Making distributional assumptions about the variables in the Causal Graph allows us to compute log-probabilities of any latent factor sample. Due to the DAG structure and the fact that all variables in $\mathcal{E}$ are independent, we know that $(X_i \indep X_j \mid Pa_i), \: \forall i=1..n, \, \forall j < i, \, X_j \not \in Pa_i$. This is easily demonstrable using d-separation \cite{pearl_causality}, since any path between non-immediately-connected nodes must pass through one of the parents of node $X_i$. Accordingly, $P(X = (x_1, \dots, x_n)) = \prod_{i=1..n} P(X_i = x_i \mid X_j = x_j, \, \forall X_j \in Pa_i)$. Each of these probability terms can be computed using the PMF (for discrete variables) or the PDF (for continuous variables) of the corresponding distribution, using the estimated parameters from the corresponding MLP network. In other words, if $\theta_i$ represents the parameter(s) of node $X_i$'s distribution, which is the output of the MLP $f_i$, $\theta_i = f_i(X_j = x_j, \, \forall X_j \in Pa_i)$. Therefore, $\log P(X = (x_1, \dots, x_n)) = \sum_{i=1..n} \log P(X_i = x_i \mid \theta_i)$. 

This tells us how to compute distribution parameters and log-likelihoods, but not how to sample. Notice that we do not use any exogenous noise in the estimation of parameters $\theta_i$. That is because the noise signal is reserved to sampling.

Following the same line as the reparametrization trick in Kingma \etal \cite{kingma_auto-encoding_2013}, we assume that each exogenous noise signal $E_i$ is a parameter-independent sample that we transform, using the parameters $\theta_i$ and an appropriate transformation function, to sample from the target distribution. Let us suppose a univariate Gaussian node $X_i \sim \mathcal{N}(\mu(Pa_i), \sigma(Pa_i))$. Then, given a noise signal $e_i$ sampled from a $\mathcal{N}(0, 1)$ prior, $x_i$ can be sampled by computing the following deterministic, differentiable expression: $x_i = \mu(Pa_i) + \sigma(Pa_i) \cdot e_i$. This trick allows us to backpropagate from a sample to its parameters, therefore allowing optimization with Gradient Descent techniques, even from the log-likelihood of the sample (as long as the PMF/PDF is differentiable).

Note that any distribution that allows such an operation, sampled from any other independent distribution (not necessarily Gaussian) can automatically be used in this framework. Bernoulli and Categorical distributions can be sampled through the use of the so-called Gumbel Trick (Papandreou \& Yuille \cite{papandreou_perturb-and-map_2011}). Implicit Reparametrization Gradients (Figurnov \etal  \cite{figurnov_implicit_2018}) offers an alternative method that allows, in particular, sampling from any distribution with a differentiable CDF, such as Truncated Univariate Normal Distributions, mixtures of reparametrizatable distributions, Gamma, Student's t or even Von Mises distributions. Hence, as long as we are able to sample differentiably from a parametric distribution, we can apply it to our DCG. Additionally, if the corresponding PMF or PDF is differentiable with respect to the distribution's parameters, we can even backpropagate through log-likelihoods of latent factor samples.

In order to train the DCG, we use Maximum Likelihood Estimation. We compute the output of each node using the samples from our dataset as parent values. If we have a sample $x = (x_1, \dots, x_n)$, we obtain the distribution parameters by running the MLP to obtain $\theta_i = \theta_i(x_j :\: \forall j,\,X_j \in Pa_i)$. Then, we can compute the log-likelihood of the whole sample and use the mean log-likelihood of a batch as a maximizing objective for optimization.

\subsection{Counterfactual Image Generator}

The next step in the process is to translate a latent factor sample into an image. However, not any image suffices. Our latent factor model might not describe an image completely, and the same vector might relate to several different examples. Therefore, we need the original image as an anchor of the generative process, so that the resulting counterfactual is both similar to the original input (preserving those factors that the causal model does not consider) and also fulfills the demands of the counterfactual latent factor vector that we created with the DCG. 

Hence, the structure of a Counterfactual Image Generator asks for an image and a counterfactual latent factor as input. The output must fulfill the latent factor (we could train an image classifier that checks if the required factors and the resulting factors match) and it should be similar to the original image. This similarity, however, cannot be simply an $L_2$ or $L_1$ loss, since the differences created by the latent factor might create significant changes in the image that these two losses cannot ignore. Therefore, one needs to identify the remaining latent factors (not present in the causal model) and then the comparison between images should only take into account these factors. 

There are several proposals in the literature that, in principle, could be accommodated to this use case, but tests revealed that their training method and/or architecture would not fulfill the requirements. CausalGAN (Kocaoglu \etal \cite{kocaoglu_causalgan:_2017}) uses a GAN architecture coupled with a Causal Controller Graph. Since the GAN only receives a random noise vector and the latent factors, it cannot be anchored to the image of study and the result can be extremely different to the original image. One could extend the original architecture to include an Encoder network, along with a reconstruction factor that assures the anchoring of the original image. However, this approach is similar to the two following ones, which did not prove effective. In any case, we do not dismiss the potential of the CausalGAN approach for this use case and might even study it further in future research.

Fader Networks \cite{lample_fader_2017} and AttGAN  \cite{he_attgan:_2019} are the two other examples we have considered. Both use an Autoencoder architecture to try to solve the problem of attribute editing in images. The former adds a critic in the hidden latent code (the output of the encoder) as a way of ensuring that the decoder learns to use the editing factors in the generated image. The latter adds the critic to the actual output, stating that the approach of Fader Networks, that try to ensure that the hidden latent code is independent of the editing factors, is too restrictive. Instead, they use a critic as adversarial loss (to generate realistic images), a classifier to adjust to the required editing factors and a reconstruction loss to anchor to the original image. 

Both approaches avoid the problem of using a reconstruction loss on the factor-altered version of the image, since they can pass the original editing values to try to reconstruct the original image. However, in that way there exists a tension between the reconstruction of the anchoring image and the generation of new samples, that, in our experiments, renders them too sensitive to the weights applied to each loss term. We argue that this could be a reason behind the fact that the quality of their images decreases significantly as more factors are considered (\ie edition on moustache, makeup and eyeglasses at the same time). It is not necessarily a problem in the image editing setting, but when trying to create counterfactual images, one needs to alter all factors at once, since any intervention can potentially affect multiple factors at the same time. As a result, the quality of the counterfactuals is vastly affected and their validity for intervention effect estimation is compromised.

Additionally, these two approaches suffer from a more significant drawback. Since they train using an Adversarial Loss, any non-realistic image is penalized and, therefore, not properly trained. This might happen with any low-probability factor configuration, as "woman with moustache" in a celebrities face dataset. Therefore, if one wants to estimate the effect of the appearance of a moustache in the image of a woman, the generator might not achieve this transformation or might start removing attributes that conform with their notion of womanhood so that the resulting image can include the moustache. This specific example is actually achieved in AttGAN, but as more attributes are considered the quality and feasibility of the edit decreases. CausalGAN, on the other hand, is prepared for this kind of edits, since its generator is attached to a trained Causal Graph that allows for such an intervention. This highlights the importance of taking into account the causal structure of the data: if an intervention is possible, the generator must learn to perform it. That is why causal blindness is a significant drawback of the previous two approaches.

In conclusion, this analysis highlights the desiderata for a Counterfactual Image Generator. As far as we know, there is no approach in the literature that accomplishes these requirements, which motivates further research on the topic.

\begin{itemize}
    \item Causal Generator: consider any configuration sampled from an arbitrary intervention, even if it is not likely in the observational data, because it might be in the intervened, counterfactual model.
    \item Reconstruction loss applied only on latent factors not present in the causal graph: any factor present in the graph must be ignored in the reconstruction loss, since it is subject to change in the edited image.
    \item Adversarial training: non-realistic but correctly classified images (with respect to the factor classifier) might appear without a critic to discard them.
\end{itemize}

As a final note, it is worth mentioning that other kinds of problems (\ie tabular data) do not need a generator at all, since the result from the DCG can be the input itself. Therefore, only problems with a separation between actual input and latent factors (\ie vision tasks) require this additional generator.

\subsection{Effect estimation}

The final part in our method is to compute the effects of any intervention on the classifier. In order to do that, we need to discuss how to compute counterfactuals in a DCG.

Causal theory normally considers exogenous noise variables ($\mathcal{E}$) as tiny perturbations in the sample due to unaccounted causes of each node. As a result, when computing counterfactuals, it is reasonable to preserve such noise signals and only change the values of the nodes affected by an intervention. That is the abduction step in the three-step process of Counterfactuals proposed by Pearl \cite{pearl_causality}. 

This is not the case in DCGs. Here, we assume that each node encodes a given distribution, and the noise signal is the mechanism to sample from that distribution. However, the actual function used for sampling from the independent noise signal to the node's distribution specified by its parameters induces an unavoidable bias in the assignment of the value. 

Consider, as an example, a Categorical variable of 3 categories determined by parameters $p=(p_1, p_2, p_3)$, with $\sum_{i=1..3} p_i = 1$. If the noise signal $e$ comes from a Uniform distribution, $e \sim \mathcal{U}(0, 1)$, one way to sample from $\textit{Categorical}(p_1, p_2, p_3)$ is to perform the operation $f(p_1, p_2, p_3, e) = 1 \cdot \mathbbm{1}_{p_1 \leq X < p_1 + p_2}(e) + 2 \cdot \mathbbm{1}_{p_1 + p_2 \leq X}(e)$, where $\mathbbm{1}_P(.)$ is the indicator function that returns a 1 when the input satisfies predicate $P$ and 0 otherwise. Notice that a value $e < p_1$, when applied to a different $\textit{Categorical}(p'_1, p'_2, p'_3)$ is biased towards low classes, making the third class less likely. If, instead, we use a different sampling function, \ie, if we split the $(0, 1)$ interval in an arbitrary number of tiny sub-intervals for each class, so that their summed volume is the corresponding $p_i$ probability, then this specific bias does not appear. Hence, the choice of sampling method can bias the results of our counterfactual for our DCG.

Consequently, instead of performing the abduction step, we just intervene on the graph and then resample all nodes who have any of the intervened nodes as ancestors. This maintains the "state of the world" for any nodes not dependent on the intervention performed but allows a different sampling affected by the new interventions on the remainder. We call this property \textit{weak-stability}, since only nodes who have none of the intervened nodes as ancestors are stable (maintain their exogenous noise values), while the rest of them are resampled. 

With this counterfactual sampling method, several different counterfactuals can appear. Each configuration is passed through the generator to obtain a new (counterfactual) image and then it can be passed through the to-be-explained classifier to obtain its prediction. The form of this prediction can be either the actual classification or its logits. The choice of which to use, or the use of both, depends on the use case. These counterfactual outcomes can then be averaged for all counterfactual images to estimate the average effect of the performed intervention on the final classifier. Additionally, one can use several weighting mechanisms on that average, either by using the log-likelihood of each counterfactual factor configuration, or by using a "proximity" measure between the counterfactual and the original factors, so that closer worlds are more relevant in the final estimation. The choice between one weighting scheme or the other also depends on the use case of the explanations.

\section{Experiments}

\begin{figure}
    \centering
    \includegraphics[width=.8\linewidth]{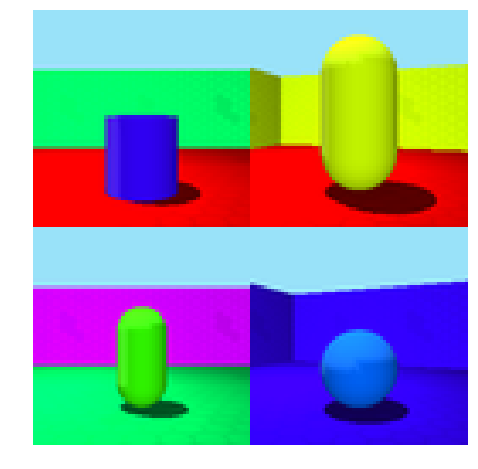}
    \caption{Samples from 3D Shapes dataset (Burgess \etal \cite{3dshapes18}).}
    \label{fig:3dshapes}
\end{figure}

In this section we detail the experiments performed to test this approach. Due to the difficulties of the Image Generator, we base our experiments on a synthetic problem, an extension to the 3D Shapes dataset (Burgess \etal \cite{3dshapes18}). This dataset consists of 480,000 synthetic images like the ones in figure \ref{fig:3dshapes}. It contains 6 different factors ({\em object, wall and floor hue, object shape and scale, and camera orientation}) and all possible combinations of these factors are included in the dataset. In this setting, the image generator can be replaced by a simple look-up in the dataset to retrieve the counterfactual image. 

In order to perform a meaningful attribution study, we design an artificial causal structure for the factors. We create a custom sampling process so that certain configurations are more likely than others. Additionally, we create two other latent factors: \textit{brightness}, which is the mean brightness of the image (created by modifying it), and \textit{type}, which is the target variable that the classifier of study will try to infer. Our objective will be to explain this classifier.

\begin{figure}
    \centering
    \includegraphics{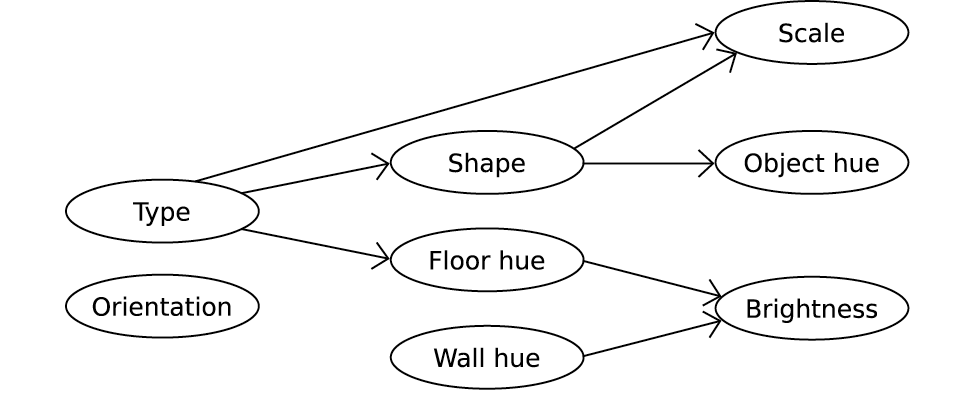}
    \caption{Causal graph for 3D Shapes dataset}
    \label{fig:3dshapes_graph}
\end{figure}

\textit{Type} is a random Bernoulli variable that is not present in the image and is not causally dependent on any other factors. It is just a coin toss, with a probability of 40\% for type 1, and 60\% for type 0. \textit{Type}, however, is used to sample the rest of the factors in a causal way. The causal graph in figure \ref{fig:3dshapes_graph} shows the relationship between the variables. \textit{Orientation} is an independent variable with 15 possible values, used to have 15 different images with the same factor configurations in the rest of the variables. \textit{Brightness} is computed as the mean of \textit{Floor hue} and \textit{Wall hue} (both considered to be numbers in $[0, 1]$) plus a small normal noise. The resulting value is applied as the mean brightness of the image to reflect the effect of the factor. Finally, this last variable is conditioned by rejection sampling, rejecting any samples that have \textit{Brightness} outside of $[0.4, 0.6]$. This means that any latent factor configurations that result in not fulfilling the \textit{Brightness} condition are automatically rejected. This is done to create a spurious correlation between \textit{Floor hue} and \textit{Wall hue} as a result of collider bias, and since \textit{Floor hue} is causally-dependent on \textit{Type}, \textit{Wall hue} is also correlated to \textit{Type} in the data, although no causal relationship exists. This artifact is used to test the fitting capabilities of the DCG in a setting where the causal graph does not explain certain non-causal correlations.

We train a DCG with all nodes except \textit{Orientation}. About distributional assumptions, \textit{Type} is considered a Bernoulli node, \textit{Shape} a 4-dimensional Categorical variable and the remaining nodes as Truncated Normal distributions, truncated to interval $[0, 1]$ except for \textit{Scale}, which is in the interval $[0.75, 1.25]$. Note that even if we know that \textit{Brightness} is restricted to $[0.4, 0.6]$, we do not apply this knowledge in the distribution of its node, to see if the graph adapts to this involuntary mismatch.

\begin{figure}
    \centering
    \includegraphics[width=\linewidth]{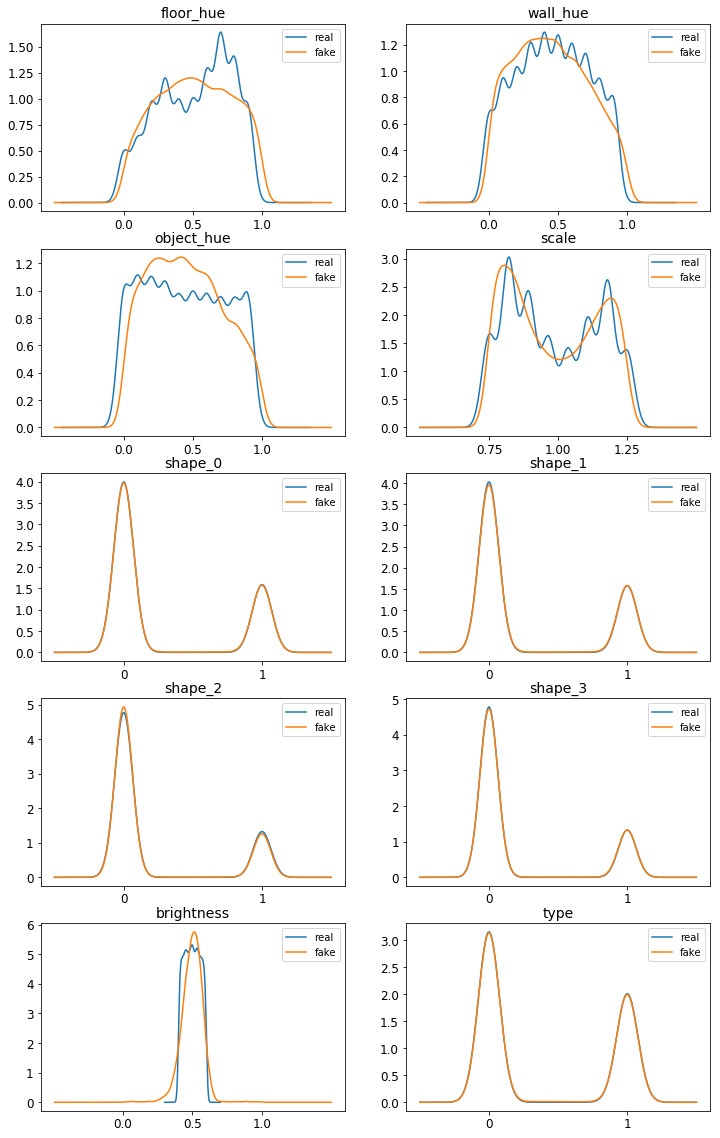}
    \caption{3DShapes KDE plots, comparing the dataset (real) and the generated (fake) distributions.}
    \label{fig:3dshapes_match}
\end{figure}

The results are displayed in figure \ref{fig:3dshapes_match}, where Kernel Density Estimation (KDE) is used in two batches of 10,000 samples, one from the original dataset, one sampled from the trained graph. As the graphs show, Bernoulli and Categorical variables are well approximated (their $p$ parameters match with the data) while continuous variables get mixed results. Note that \textit{hue} and \textit{Scale} are defined by a discrete number of possible values in their respective intervals; this accounts for the waving pattern in the KDE estimation that our distributions cannot mimic. \textit{Brightness}, due to its implicit conditioning to $[0.4, 0.6]$, is not as well approximated as it should, but the graph manages to stay in the interval nonetheless. Finally, hue variables are the worst approximations, due to their cyclic nature ($\textit{hue} = 1$ is the same as $\textit{hue} = 0$) and their discrete values. By using a cyclic distribution, such as the Von Mises distribution, these results could improve further, but for our purposes, this approximation is enough.

\begin{figure}
    \centering
    \includegraphics[width=\linewidth]{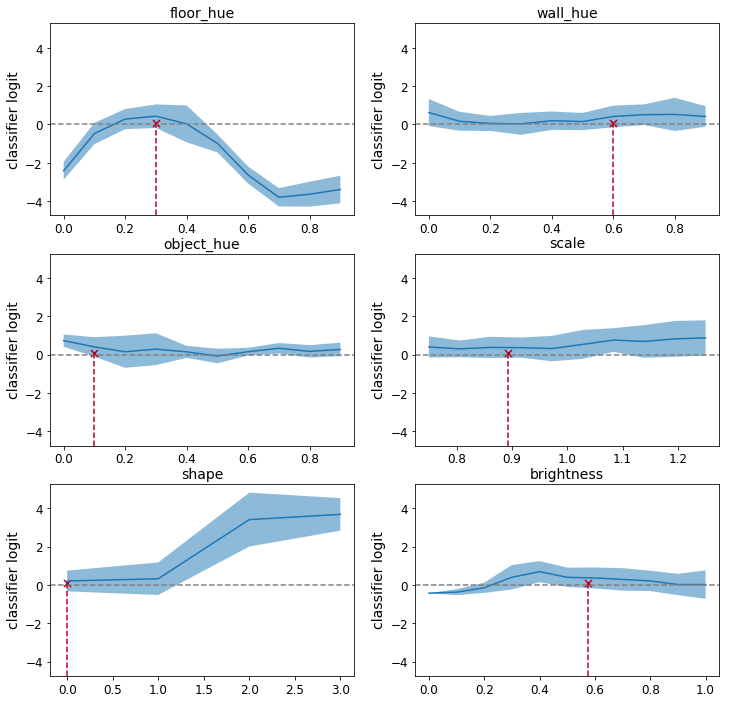}
    \caption{3DShapes 1-dimensional intervention effect estimation. The red cross signals the real value for that factor, and the corresponding classifier logit.}
    \label{fig:3dshapes_effects1d}
\end{figure}

In terms of computing the effects of interventions, we include an explanation example in figure \ref{fig:3dshapes_effects1d}. For a given image with a wrong prediction (predicted as $\mathit{Type}=1$ when it actually was $\mathit{Type}=0$), we want to explain why the classifier made a mistake. The x-axis shows the different values that each latent factor can take, while the y-axis shows the classifier's pre-sigmoid score, the logits. The horizontal line $y = 0$ corresponds to the decision boundary, so values below the line are correct predictions. Given the original image latent factor values, shown as the red cross in each plot, we compute 100 counterfactual samples for each possible intervention (each possible value) of all variables. These samples allow us to compute the average classifier prediction for counterfactuals based on the original sample and the specified intervention, but also the standard deviation of these predictions, with which (assuming a Gaussian distribution of the effects) we can compute a Confidence Interval, shown as the blue tinted area. 

Thanks to this visualization, we can see that the only factor that effectively changes from the wrong class to the right one is \textit{Floor hue}. This tells us that, when only considering 1-factor interventions, the reason for the wrong prediction to this particular image is the fact that \textit{Floor hue} had a value in $[0.2, 0.4]$. Had it been in $[0.6, 0.9]$, the classifier would have made the right prediction. Therefore, we can say that \textit{Floor hue} caused the error. 

It is important to note that not all values are feasible. In the case of brightness, we know that it can only take values in $[0.4, 0.6]$, so the effects outside this region might be meaningless depending on the application. One could also want to study 2-factor effects, by just computing any possible combination of the two factors. We do not include this experiment because of space restrictions.

\section{Conclusion}

In this paper, we proposed an approach for Visual Model Explanation by focusing on the causal relationships between the latent factors in the data. With the use of Distributional Causal Graphs, such explanations could take into account the likelihood of the counterfactuals along with their effects, which could be beneficial for applications that look for feasible interventions to achieve desired results. 

The only remaining piece is the training of a proper Counterfactual Image Generator to plug in our method. We have discussed and identified the problems that current state of the art image generators face when dealing with this problem. Thanks to that, we could establish the desiderata for such a model, which we hope will motive further research in the topic. 

\section*{{\small Acknowledgements}}
\noindent {\small This work has been partially funded by projects TIN2015-66951-C2, RTI2018-095232-B-C21 (MINECO/FEDER) and 2017.SGR.1742 (Generalitat de Catalunya).}

{\small
\bibliographystyle{ieee}
\bibliography{Explaining_Visual_Models_by_Causal_Attribution}

\begin{thebibliography}{10}\itemsep=-1pt

\bibitem{ancona_towards_2018}
M.~Ancona, E.~Ceolini, C.~Oztireli, and M.~Gross.
\newblock Towards better understanding of gradient-based attribution methods
  for {Deep} {Neural} {Networks}.
\newblock In {\em 6th {International} {Conference} on {Learning}
  {Representations} (ICLR)}, 2018.

\bibitem{3dshapes18}
C.~Burgess and H.~Kim.
\newblock 3d shapes dataset.
\newblock https://github.com/deepmind/3dshapes-dataset/, 2018.

\bibitem{chang_explaining_2018}
C.-H. Chang, E.~Creager, A.~Goldenberg, and D.~Duvenaud.
\newblock Explaining image classifiers by counterfactual generation.
\newblock In {\em International Conference on Learning Representations (ICLR)},
  2019.

\bibitem{denton_detecting_2019}
E.~Denton, B.~Hutchinson, M.~Mitchell, and T.~Gebru.
\newblock Detecting {Bias} with {Generative} {Counterfactual} {Face}
  {Attribute} {Augmentation}.
\newblock {\em arXiv:1906.06439 [cs, stat]}, Jun. 2019.

\bibitem{figurnov_implicit_2018}
M.~Figurnov, S.~Mohamed, and A.~Mnih.
\newblock Implicit {Reparameterization} {Gradients}.
\newblock In {\em Advances in {Neural} {Information} {Processing} {Systems}
  31}, pages 441--452. 2018.

\bibitem{goodfellow_generative_2014}
I.~Goodfellow, J.~Pouget-Abadie, M.~Mirza, B.~Xu, D.~Warde-Farley, S.~Ozair,
  A.~Courville, and Y.~Bengio.
\newblock Generative {Adversarial} {Nets}.
\newblock In {\em Advances in {Neural} {Information} {Processing} {Systems}
  27}, pages 2672--2680. 2014.

\bibitem{goyal_counterfactual_2019}
Y.~Goyal, Z.~Wu, J.~Ernst, D.~Batra, D.~Parikh, and S.~Lee.
\newblock Counterfactual visual explanations.
\newblock In {\em Proceedings of the 36th International Conference on Machine
  Learning}, volume~97, pages 2376--2384, June 2019.

\bibitem{he_attgan:_2019}
Z.~He, W.~Zuo, M.~Kan, S.~Shan, S.~Shan, and X.~Chen.
\newblock {AttGAN}: {Facial} {Attribute} {Editing} by {Only} {Changing} {What}
  {You} {Want}.
\newblock {\em IEEE Transactions on Image Processing}, 2019.

\bibitem{kim_interpretability_2017}
B.~Kim, M.~Wattenberg, J.~Gilmer, C.~Cai, J.~Wexler, F.~Viegas, and R.~sayres.
\newblock Interpretability beyond feature attribution: Quantitative testing
  with concept activation vectors ({TCAV}).
\newblock In {\em Proceedings of the 35th International Conference on Machine
  Learning}, volume~80, pages 2668--2677, Jul. 2018.

\bibitem{kingma_auto-encoding_2013}
D.~P. Kingma and M.~Welling.
\newblock {Auto-encoding variational Bayes}.
\newblock In {\em Proceedings of the International Conference on Learning
  Representations (ICLR)}, 2014.

\bibitem{kocaoglu_causalgan:_2017}
M.~Kocaoglu, C.~Snyder, A.~G. Dimakis, and S.~Vishwanath.
\newblock Causal{GAN}: Learning causal implicit generative models with
  adversarial training.
\newblock In {\em International Conference on Learning Representations (ICLR)},
  2018.

\bibitem{lample_fader_2017}
G.~Lample, N.~Zeghidour, N.~Usunier, A.~Bordes, L.~Denoyer, and M.~A. Ranzato.
\newblock Fader {Networks}: {Manipulating} {Images} by {Sliding} {Attributes}.
\newblock In {\em Advances in {Neural} {Information} {Processing} {Systems}
  30}, pages 5967--5976. 2017.

\bibitem{lipton_mythos_2016}
Z.~C. Lipton.
\newblock The mythos of model interpretability.
\newblock {\em Communications of the ACM}, 61(10):36--43, 2018.

\bibitem{mothilal_explaining_2019}
R.~K. Mothilal, A.~Sharma, and C.~Tan.
\newblock Explaining {Machine} {Learning} {Classifiers} through {Diverse}
  {Counterfactual} {Explanations}.
\newblock {\em arXiv:1905.07697 [cs, stat]}, May 2019.

\bibitem{papandreou_perturb-and-map_2011}
G.~Papandreou and A.~L. Yuille.
\newblock Perturb-and-{MAP} random fields: {Using} discrete optimization to
  learn and sample from energy models.
\newblock In {\em 2011 {International} {Conference} on {Computer} {Vision}},
  pages 193--200, Nov. 2011.

\bibitem{pearl_causality}
J.~Pearl.
\newblock {\em Causality: {Models}, {Reasoning} and {Inference}}.
\newblock Cambridge University Press, New York, NY, USA, 2nd edition, 2009.

\bibitem{sundararajan_gradients_2016}
M.~Sundararajan, A.~Taly, and Q.~Yan.
\newblock Gradients of {Counterfactuals}.
\newblock {\em arXiv:1611.02639 [cs]}, Nov. 2016.

\bibitem{wachter_counterfactual_2017}
S.~Wachter, B.~Mittelstadt, and C.~Russell.
\newblock Counterfactual {Explanations} without {Opening} the {Black} {Box}:
  {Automated} {Decisions} and the {GPDR}.
\newblock {\em Harvard Journal of Law \& Technology (Harvard JOLT)}, 31:841,
  2017.

\bibitem{zeiler_visualizing_2014}
M.~D. Zeiler and R.~Fergus.
\newblock Visualizing and {Understanding} {Convolutional} {Networks}.
\newblock In {\em Computer {Vision} {ECCV} 2014}, Lecture {Notes} in {Computer}
  {Science}, pages 818--833. Springer International Publishing, 2014.

\end{thebibliography}
}

\end{document}